# Automatic Diagnosis of Abnormal Tumor Region from Brain Computed Tomography Images Using Wavelet Based Statistical Texture Features

A. Padma[1] and Dr.R. Sukanesh[2]

[1] Dept of Information Technology, Velammal College of Engg and Tech, Madurai,India

Email id: giri_padma2000@yahoo.com

[2] Prof of Electronics and Communication Eng, Thiagarajar College of Engg,Madurai,India

Email id : rshece@tce.edu

## Abstract

The research work presented in this paper is to achieve the tissue classification and automatically diagnosis the abnormal tumor region present in Computed Tomography (CT) images using the wavelet based statistical texture analysis method. Comparative studies of texture analysis method are performed for the proposed wavelet based texture analysis method and Spatial Gray Level Dependence Method (SGLDM). Our proposed system consists of four phases i) Discrete Wavelet Decomposition (ii) Feature extraction (iii) Feature selection (iv) Analysis of extracted texture features by classifier. A wavelet based statistical texture feature set is derived from normal and tumor regions. Genetic Algorithm (GA) is used to select the optimal texture features from the set of extracted texture features. We construct the Support Vector Machine (SVM) based classifier and evaluate the performance of classifier by comparing the classification results of the SVM based classifier with the Back Propagation Neural network classifier(BPN). The results of Support Vector Machine (SVM), BPN classifiers for the texture analysis methods are evaluated using Receiver Operating Characteristic (ROC) analysis. Experimental results show that the classification accuracy of SVM is 96% for 10 fold cross validation method. The system has been tested with a number of real Computed Tomography brain images and has achieved satisfactory results.

# 1. Introduction

In recent years, medical CT Images have been applied in clinical diagnosis widely. That can assist physicians to detect and locate Pathological changes with more accuracy. Computed Tomography images can be distinguished for different tissues according to their different gray levels. The images, if processed appropriately can offer a wealth of information which is significant to assist doctors in medical diagnosis. A lot of research efforts have been directed towards the field of medical image analysis with the aim to assist in diagnosis and clinical studies [1]. Pathologies are clearly identified using automated CAD system [2]. It also helps the radiologist in analyzing the digital images to bring out the possible outcomes of the diseases. The medical images are obtained from different imaging systems such as MRI scan, CT scan, Ultra sound B scans. The computerized tomography has been found to be the most reliable method for early detection of tumors because this modality is the mostly used in radio therapy planning for two main reasons. The first reason is that scanner images contain anatomical information which offers the possibility to plan the direction and the entry points of radio therapy rays which have to target only the tumor region and to avoid other organs. The second reason is that CT scan images are obtained using rays, which is same principle as radio therapy. This is very important because the intensity of radio therapy rays have been computed from the scanned image. Advantages of using CT include good detection of calcification, hemorrhage and bony detail plus lower cost, short imaging times and widespread availability. The situations include patient who are too large for MRI scanner, claustrophobic patients, patients with metallic or electrical implant and patients unable to remain motionless for the duration of the examination due to age, pain or medical condition. For these reasons, this study aims to explore methods for classifying and segmenting brain CT images. Image segmentation is the process of partitioning a digital image into set of pixels. Accurate, fast and reproducible image segmentation techniques are required in various applications. The results of the segmentation are significant for classification and analysis purposes. The limitations for CT scanning of head images are due to partial volume effects which affect the edges produce low brain tissue contrast and yield different objects within the same range of intensity. All these limitations have made the segmentation more difficult. Therefore, the challenges for automatic segmentation of the CT brain images have many different approaches. The segmentation techniques proposed by Nathali Richard et al and Zhang et al [3][4] include statistical pattern recognition techniques. Kaiping et al [5] introduced the effective Particle Swarm optimization algorithm to segment the brain images into Cerobro spinal fluid (CSF) and suspicious abnormal regions but without the annotation of the abnormal regions. Dubravko et al and Matesin et al [6] [7] proposed the rule based approach to label the abnormal regions such as calcification, hemorrhage and stroke lesion. Ruthmann.et al [8] proposed to segment Cerobro spinal fluid from computed tomography images using local thresholding technique based on maximum entropy principle. Luncaric et al proposed [9] to segment CT images into background, skull, brain, ICH, calcifications by using a combination of K means clustering and neural networks. Tong et al proposed [10] to segment CT images into CSF,brain matter and detection of abnormal regions using unsupervised clustering of two stages. Clark et al [11] proposed to segment the brain tumor automatically using knowledge based techniques. From the above literature survey shows that intensity based statistical features are the straightest forward and have been widely used, but due to the complexity of the pathology in human brain and the high quality required by clinical diagnosis, only intensity features cannot achieve acceptable result. In such applications, segmentation based on textural feature methods gives more reliable results. Therefore texture based analysis has been presented for tumor segmentation such as SGLDM method and wavelet based texture features are used and achieve promising results.

Based on the above literature, better classification accuracy can be achieved using wavelet based statistical texture features. In this paper, the authors would like to propose a wavelet based statistical texture analysis method to segment the soft tissues and automatically diagnosis abnormal tumor region from brain CT images. The proposed method is illustrated in Figure 1. This system uses the classifiers SVM [12], BPN [13] to classify and segment the abnormal tumor region from brain CT images and gives relatively good segmentation results as compared to the literature discussed above.

In our work, first by applying 2 level Discrete Wavelet Transform(DWT),the image is represented by one approximation and three detail sub bands and the co-occurrence matrix[14,15] is derived for detail sub bands. Then from these co-occurrence matrices, the statistical texture features are extracted using the SGLDM method.. The extracted texture features are optimized by Genetic Algorithm(GA)[16] for improving the classification accuracy and reducing the overall complexity. The optimal texture features are fed to the SVM,BPN classifiers to classify and segment the abnormal tumor region from brain CT images.

## 2. Materials and methods

Most classification techniques offer intensity based statistical features. However in our approach, we adopt wavelet based statistical texture features to classify and segment the abnormal tumor region. The proposed system is divided into 4 phases (a) Discrete Wavelet Decomposition (b) Feature extraction (c) Feature selection (d) Classification and Evaluation. In the proposed system for feature extraction, we discovered two methods which are wavelet based statistical texture feature extraction method, SGLDM method without wavelet transform. Firstly the two level wavelet decomposition is performed to decompose the image into one approximation and three detail images and the co-occurrence matrix is derived for $2^{nd}$ level detail images. Then from these co-occurrence matrices, the statistical texture features are extracted using the SGLDM method. Once all the features are extracted, then for feature selection, we use Genetic Algorithm(GA) to select the optimal texture features. The selected optimal texture features are given as input to the SVM ,BPN classifiers to classify and segment the abnormal tumor region from brain CT images.

## 2.1 Discrete Wavelet Decomposition

A two level wavelet decomposition of region of interest(ROI) is performed which results in four sub bands. Daubechies wavelet filter of order two is used. In 2D wavelet decomposition[17] the image is represented by one approximation and three detail images representing the low and high frequency contents image respectively. The approximation can be further to produce one approximation and three detail images at the next level of decomposition, wavelet decomposition process is shown in Figure 1. A1 and A2 represent the wavelet approximations at $1^{st}$ and $2^{nd}$ level respectively, and are low frequency part of the images. H1,V1,D1,H2,V2,D2 represent the details of horizontal, vertical and diagonal directions at $1^{st}$ and $2^{nd}$ level respectively, and are high frequency part of the images.

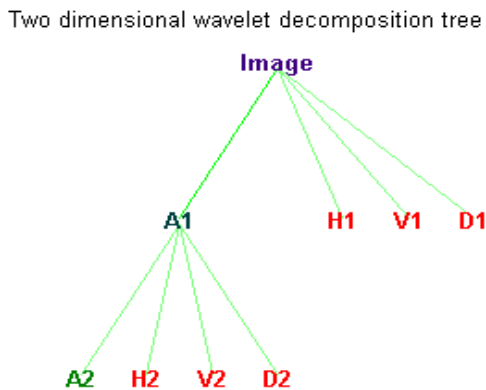

Figure 1. Two level discrete wavelet decomposition.

Among the high frequency sub bands, the one whose histogram presents the maximum variance is the sub band that represents the clearest appearance of the changes between the different textures. The textures features are extracted from these high frequency sub bands are useful to classify and segment the abnormal tumor region from brain CT images.

## 2.2 Feature extraction

Texture analysis is a quantitative method that can be used to quantify and detect structural abnormalities in different tissues .As the tissues present in brain are difficult to classify using shape or intensity level of information, the texture feature extraction is founded to be very important for further classification. The purpose of feature extraction is to reduce original data set by measuring certain features that distinguish one region of interest from another. The analysis and characterization of textures present in the medical images can be done by using wavelet based statistical feature extraction method. Each sub image is taken from top left corner of the original image is decomposed using two level DWT and co-occurrence matrices are derived for detail or high frequency sub bands(i.e.,H2,V2,D2 sub bands). Then from these co-occurrence matrices ,the Wavelet Co-occurrence Texture features (WCT)are computed.

**Algorithm for feature extraction is as follows**

- Obtain the sub-image blocks, starting from the top left corner.
- Decompose sub-image blocks using 2-D DWT.
- Derive SGLDM or Co-occurrence matrices for detail sub-bands of DWT with 1 for distance and **0**,45,90 and 135 degrees for θ and averaged.
- From these co-occurrence matrices, the following nine Haralick texture features [ 18] called wavelet Co-occurrence Texture features(WCT) are extracted.

Then the feature values are normalized by subtracting minimum value and dividing by maximum value minus minimum value. Maximum and minimum values are calculated based on the training data set. In the data set, if the feature value is less than the minimum value, it is set to minimum value. If the feature value is greater than the maximum value, it is set to maximum value. Normalized feature values are then optimized by feature selection algorithm. Table 1 Shows the WCT features extracted using SGLDM method.

Table 1. WCT Features extracted using SGLDM method

| SI.No | Second order WCT features |
|---|---|
| 1 | Entropy-ENT   (Measure the disorder of an image) |
| 2 | Energy- ENE    ( Measure the  textural uniformity ) |
| 3 | Contrast-CON     (Measure the local contrast in an image) |
| 4 | Sum Average-SA (Measure the average of the gray level within an image) |
| 5 | Variance –VAR   (Measure   the  heterogeneity of an image) |
| 6 | Correlation-COR (Measure a correlation of pixel pairs on gray levels) |
| 7 | Max probability-MP (Determine the most prominent pixel pair in an image) |
| 8 | Inverse Difference Moment - IDM   (Measure the  homogeneity of an image) |
| 9 | Cluster tendency-CT    (Measure the grouping  of pixels that have  similar |

**2.3 Feature selection**

Feature selection is the process of choosing subset of features relevant to particular application and improves classification by searching for the best feature subset, from the fixed set of original features according to a given feature evaluation criterion(i.e., classification accuracy). Optimized feature selection reduces data dimensionalities and computational time and increase the classification accuracy. The feature selection problem involves the selection of a subset of `d' features from a total of `D' features, based on a given optimization criterion. The D features are denoted uniquely by distinct numbers from 1 to D, so that the total set of D features can be written as S = { 1, 2, . . . , D }. X denotes the subset of selected features and Y denotes the set of remaining features. So,

S = X U Y at any time. J(X) denotes a function evaluating the performance of X. J depends on the particular application. Here J(X) denotes the classification performance of classifying and segmenting abnormal tumor region from brain CT images using the set of features in X. In this work, Genetic Algorithm (GA technique is used.

**Genetic algorithm:**

We consider the standard GA to begin by randomly creating its initial population. Solutions are combined via a crossover operator to produce offspring, thus expanding the current population of solutions. The individuals in the population are then evaluated via a fitness function, and the less fit individuals are eliminated to return the population to its original size. The process of crossover, evaluation, and selection is repeated for a predetermined number of generations or until a satisfactory solution has been found. A mutation operator is generally applied to each generation in order to increase variation. In the feature selection formulation of the genetic algorithm ,individuals are composed of bit strings: a 1 in bit position indicates that feature should be selected; 0 indicates this feature should not be selected. As an example bit string 00101000 means the $3^{rd}$ and $5^{th}$ features are selected. That is the chromosome represents X={3,5} and Y={1,2,4,6,7,8}. Fitness function for given bit string X is defined as

$$\text{Fitness}(X) = J(X) - \text{penalty}(X) \qquad [1]$$

Where X is the corresponding feature subset , and penalty(X) = w * (|X| -d) with a penalty coefficient w. The size value d is taken as a constraint and a penalty is imposed on chromosomes breaking this constraint. The chromosome selection for the next generation is done on the basis of fitness. The fitness value decides whether the chromosome is good or bad in a population. The selection mechanism should ensure that fitter chromosomes have a higher probability survival. So, the design adopts the rank-based roulette-wheel selection scheme. If the mutated chromosome is superior to both parents, it replaces the similar parent. If it is in between the two parents, it replaces the inferior parent; otherwise, the most inferior chromosome in the population is replaced. The selected optimal feature set based on the test data set is used to train the SVM,BPN classifiers to classify and segment the abnormal tumor region from brain CT images. Table 2 shows the Best chromosomes selected (i.e., best features) using Genetic Algorithm(GA) during the execution.

Table 2  Best chromosomes selected by GA

| SI-NO | Feature set | Classification accuracy |
|---|---|---|
| 1 | IDM,ENT, ENE, VAR,CON | 95% |
| 2 | IDM,CON,ENE, MP, VAR | 95% |
| 3 | ENT,IDM,VAR, IDM, CT | 96% |
| 4 | IDM,ENT, CT, ENE,CON | 95% |
| 5 | CON, IDM,VAR,ENT,ENE | 96% |
| 6 | ENT,SA, IDM, ENE,VAR | 96% |
| 7 | VAR,ENT, ENE, SA,IDM | 95.5% |
| 8 | ENT,CON,ENE,VAR,IDM | 96% |
| 9 | IDM,ENT, CT,CON, VAR | 96% |
| 10 | ENE,ENT, MP,CON, COR | 95% |

The texture features Energy(ENE), Entropy(ENT), Variance(VAR), and Inverse Difference Moment(IDM) are present in most of the feature vectors or feature set selected by GA. The features such as CT,CON,SA,MP,COR which are least significant. The classification accuracy of 96% is obtained with four of the available 9 features using GA. Therefore it is possible to classify and segment abnormal tumor region from brain CT images.

### 2.4  SVM classifier

Classification is the process where a given test sample is assigned a class on the basis of knowledge gained by the classifier during training. Support Vector Machine(SVM) performs the robust non-linear classification with kernel trick. SVM is independent of the dimensionality of the feature space and that the results obtained are very accurate. It outperforms other classifiers even with small numbers of available training samples. SVM is a supervised learning method and is used for one class and n class classification problems. It combines linear algorithms with linear or non-linear kernel functions that make it a powerful tool in the machine learning community with applications such as data mining and medical imaging applications. To apply SVM into non linear data distributions, the data can be implicitly transformed to a high dimensional feature space where a separation might become possible. For a binary classification given a set of separable data set with N samples $X = \{X_i\}$, $i = 1, 2 \ldots N$, labeled as $Y_i = \pm 1$. It may be difficult to separate these 2 classes in the input space directly. Thus they are mapped into a higher dimensional feature space by $X' = f(x)$.

The decision function can be expressed as

$$f(x) = W.x + \rho \qquad [2]$$

Where $W.x + P = 0$ is a set of hyper planes to separate the two classes in the new feature space. Therefore for all the correctly classified data,

$$Y_i f(x) = Y_i (W.x + \rho) > 0, \quad i = 1, 2 \ldots N \qquad [3]$$

By scaling W and ρ properly, we can have f(x) = W.x + ρ = 1 for those data labeled as +1 closes to the optimal hyper plane and f(x) = W.x + ρ = -1 for all the data labeled as -1 closes to the optimal hyper plane. In order to maximize the margin the following problem needs to be solved.

$$\text{Min } (\|W\|^2/2)$$

Subject to $Y_i f(x) = Y_i (W.x + \rho) \geq 1$, i = 1, 2 ….. N         [4]

It is a quadratic programming problem to maximize the margins which can be solved by sequential minimization optimization. After optimization, the optimal separating hyper plane can be expressed as

$$f(x) = \sum_{i=1}^{N} \alpha_i Y_i K(x_i, x) + \rho \qquad [5]$$

Where K(.) is a kernel function, ρ is a bias, α is the solutions of the quadratic programming problem to find maximum margin. When α is non zero, are called support vectors, which are either on or near separating hyper plane. The decision boundary (i.e.) the separating hyper plane whose decision values f(x) approach zero, compared with the support vectors, the decision values of positive samples have larger positive values and those of negative samples have larger negative values. Therefore the magnitude of the decision value can also be regarded as the confidence of classifier. The larger the magnitude of f(x), the more confidence of the classification by choosing Gaussian kernel function

$$K(x,y) = e^{-\gamma \|x-y\|^2} \qquad [6]$$

Where the value of γ was chosen to be 1 and has good performance for the following two reasons. First reason is the Gaussian model has only one parameter and it is easy to construct the Gaussian SVM classifier compared to polynomial model which has multiple parameters. Second reason is there is less limitation in using Gaussian kernel function due to nonlinear mapping in higher dimensional space.

## 2.5 Detection of abnormal tumor region

Detection is important in selecting the sub band of the image to be decomposed. The process is done by applying the 2 level 2D DWT, the image is decomposed into four sub bands. After decomposition, SGLDM or Co-occurrence matrices is derived on detail sub bands. From these co-occurrence matrices, WCT features are extracted as given in the feature extraction algorithm and the optimal texture feature set is selected by GA based on the classification performance of SVM,BPN classifiers. From the experiments conducted for feature selection, it is found that the optimal feature set which gives good classification performance are the second order WCT features like energy entropy, variance, and inverse difference moment. The four texture features from detail sub bands form the feature vectors or feature set. These feature vectors are given as input to the SVM,BPN classifiers to classify and segment the abnormal tumor region. Efficiency or accuracy of the classifiers for each texture analysis methods are analyzed based on the error rate (i.e.) All tests could have an error rate. This error rate can be described by the terms true and false positive and true and false negative as follows:

**True Positive (TP):** The test result is positive in the presence of the clinical abnormality.

**True Negative (TN):** The test result is negative in the absence of the clinical abnormality.

**False Positive (FP):** The test result is positive in the absence of the clinical abnormality.

**False Negative (FN):** The test result is negative in the presence of the clinical abnormality.

Based on the above terms to construct the contingency table.

Table 3 Contingency table of classifier performance.

| Actual Group | Predicted group | |
|---|---|---|
| | Normal | Abnormal |
| Normal | TN | FP |
| Abnormal | FN | TP |

Sensitivity = TP / (TP + FN)   Specificity = TN / (FP + TN)

Accuracy  =  (TP+TN)/(TP + TN + FN + FP)

Sensitivity measures the ability of the method to identify abnormal cases. Specificity measures the ability of the method to identify normal cases. Correct classification rate or accuracy is the proportion of correct classifications to the total number of classification tests. The SVM,BPN classifiers were tested by using Leave one out cross validation method. Leave one out cross validation can be used as a method to estimate the classifier performance in unbiased manner. Here each iteration, one data set is left out and the classifier is trained using the rest and the testing applied to the left out data set. This procedure is repeated such that each data set is left out once. Classification accuracy is calculated by taking the average number of all the iterations. To evaluate the classification accuracy of classifiers, the 10 fold cross validation is done on the data set collected from 100 images. In this method, the images are divided into 10 sets each consisting of 5 normal images and 5 abnormal images. Then 9 sets are used for training and one set is used for testing. Hence 10 iterations are done. s. For eg. In the first iteration 2 images and another two iterations 1 image are wrongly classified. For the remaining 7 iterations all are correctly classified. Hence the cross validation accuracy obtained as 96/100 which is equal to 96%. The results show that, if the number of representative samples increased, we get good classification accuracy for 10 fold cross validation method. Other statistical method known as Receiver Operating Characteristics (ROC) analysis [19] is also used to analyze the experimental results of all the classifiers. The ROC curve is a graphical representation of sensitivity versus specificity as a threshold parameter is varied. The Area under ROC Curve (AUC) has been used to determine the overall classification accuracy. By calculating AUC, we can measure the class discrimination capability of a specific classifier. An area of above 0.5 represents a perfect test while an area of less than or equal to 0.5 represents worthless test. The larger the area (the higher AUC value) means higher the classification performance. In this research, the ROC analysis and accuracy are used to measure the performance of the classifiers and texture analysis methods.

**3.Results**

Our proposed method is implemented on real human brain CT dataset based on proposed flow diagram as shown in Figure 1. Figure 2(a) represents the original CT brain image and Figure.2(b) represents abnormal tumor (benign) brain CT image and Figure. 2(c) represents the abnormal tumor region(malignant) brain CT image. The input data set consists of 100 images: 40 images are normal, 60 images are abnormal. For each texture analysis method, input data set is partitioned into training and test sets which are classified using SVM,BPN classifiers. This section describes the wavelet based texture analysis method of classifying and segmenting abnormal tumor region of CT images.

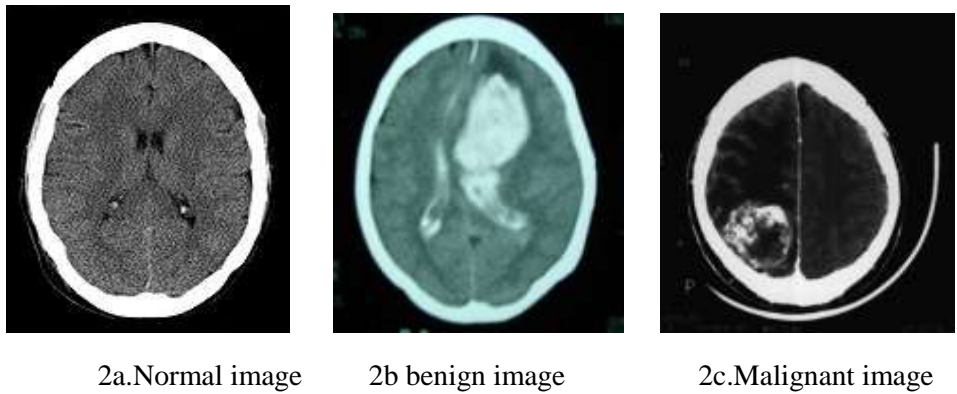

    2a.Normal image    2b benign image    2c.Malignant image

Figure 2.Example of CT Normal and tumor images

BPN classifier performance was analyzed based on the experiments with the data set of 100 images. In BPN classifier ,more than one hidden layer may be beneficial for some applications, but one hidden layer is sufficient if enough hidden neurons are used The number of hidden nodes used in BPN is 25 (i.e.) 2*number of nodes in the input layer +1, number of nodes in the input layer is 12, so 2*12+1=25, because 12 features are given as input to the classifier (Four features such as energy, entropy, variance, inverse difference moment are extracted from detail sub bands. So ( 3*4=12). For the best performance of BPN ,the proper number of nodes in the hidden layer is selected through trial and error method based on number of epochs needed to train the network .It is observed that the network performed well with 25 hidden nodes. The learning rate for input and hidden layer is 0.4, moment=0.2 and the error allowed is 0.01. Binary sigmoid function is used. After successful training with 450 epochs the satisfactory results are obtained. In BPN, the initial weights are randomly selected from [-0.5, 0.5]. The selected optimal texture features are given as input to the BPN classifier.

Feature selection is carried out using GA. There are 9 features are extracted from detail sub bands. So totally 9*3= 27 features are extracted. The next step is to determine the relevance of each selected feature to the process of classifying and segmenting abnormal tumor region. During the evaluation process by using GA, some features may be selected many times as the number of generation increases. If the feature was selected more times that feature was given as more important in the feature selection. The number of times the features selected was energy, entropy, variance and inverse difference moment. The parameter set for the GA algorithm is as follows: Population size is 30; Cross Over probability is 1.0; Mutation rate is 0.1; Penalty coefficient is 0.5 and stopping condition is 100 generations . Results show that, if the number of sample images increased, we get good classification accuracy for the 10 fold cross validation method .

A comparative study of the classification accuracy is performed for both wavelet based texture analysis method and Spatial Gray Level Dependency Matrix method. Table 5 shows the classification performances of the SVM classifier with different kernel functions.

Table 4.Classification results of SVM with different kernel functions

| Kernel used | No of images | Training | | Testing | | Images misclassified classified | Accuracy |
|---|---|---|---|---|---|---|---|
| | | N | A | N | A | | |
| Linear | 120 | 50 | 70 | 50 | 64 | 6 | 96.6% |
| Polynomial | 120 | 50 | 70 | 50 | 66 | 4 | 97% |
| Gaussian | 120 | 50 | 70 | 50 | 67 | 3 | 97.5% |

N-Normal image, A-Abnormal image

The accuracy of SVM with Gaussian kernel function and SVM with polynomial kernel function and SVM with linear function are 95%, 95.4%.96% respectively for same training and testing data sets .The accuracy of SVM with Gaussian kernel function is high while compared to SVM classifier with linear and polynomial functions.

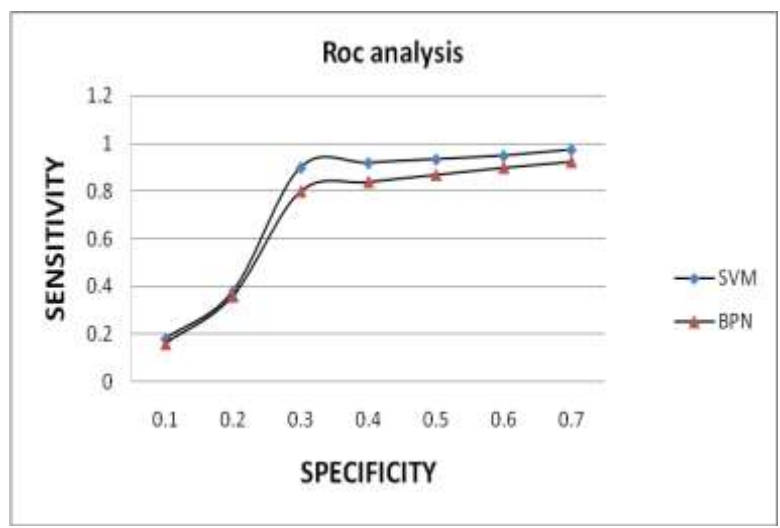

Figure 3 Roc analysis curves of classifiers in wavelet domain

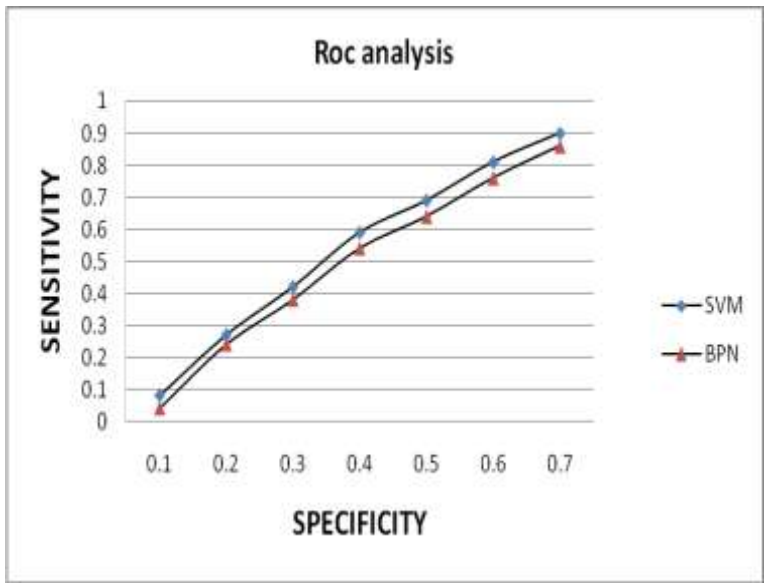

Figure 4.Roc analysis curve of classifiers in gray level domain

Figure 3 and Figure 4 shows the Roc analysis curve of classifiers in wavelet domain and gray level domain. From the ROC analysis also ,the Area under the ROC Curve of SVM classifier in wavelet domain , gray level domain are 0.96,0.92 . The Area under the ROC Curve of BPN classifier in wavelet domain , gray level domain are 0.91,0.89. To justify the choice of wavelet domain for the same data set without applying wavelet transform, in the gray level domain, the four texture features are extracted and the performance of the SVM classifier is shown in Table 6 and the performance of BPN classifier is shown in Table 7. The accuracy of SVM, BPN classifiers in wavelet domain are 96%,92% and in gray level domain are 91%, 89% respectively

Table 5 Classification performances of SVM classifier for 100 images

| Parameter used | Wavelet domain | Gray level domain |
|---|---|---|
| TP | 49 | 47 |
| TN | 47 | 45 |
| FP | 2 | 5 |
| FN | 2 | 3 |
| Sensitivity in % | 96.07% | 94% |
| Specificity in % | 95.91% | 90% |
| Accuracy in % | 96% | 92% |

Table 6 Classification performances of BPN classifier for 100 images

| Parameter used | Wavelet domain | Gray level domain |
|---|---|---|
| TP | 46 | 45 |
| TN | 45 | 44 |
| FP | 5 | 7 |
| FN | 4 | 4 |
| Sensitivity in % | 92% | 91.8% |
| Specificity in % | 90% | 86.2% |
| Accuracy in % | 91% | 89% |

Table 7 shows classification performances of our proposed technique and the SGLDM method. The classification accuracy of our proposed method is 96% which is high while compared with SGLDM method.

Table 7 Classification accuracy of proposed technique

| SI-No | Technique | Classification Accuracy |
|---|---|---|
| 1 | WT+SGLDM+GA+SVM | 96% |
| 2 | SGLDM+GA+SVM | 92% |
| 3 | WT+SGLDM+GA+BPN | 91% |
| 4 | SGLDM+GA+BPN | 89% |

## 4. Conclusions

As a conclusion, we have presented a method for wavelet based texture feature extraction and selecting the optimal texture features using GA , and evaluated the SVM, BPN classifiers to classify and segment the abnormal tumor region. The algorithm has been designed based on the concept of different types of brain soft tissues have different textural features. This method effectively works well for detection of abnormal tumor region with high sensitivity, specificity and accuracy. The classification accuracy of the SVM,BPN classifiers using wavelet based texture analysis method and SGLDM method without wavelet transform to classify and segment the abnormal tumor region are 96%,92% ,91%,89% for 10 fold cross validation method. From the ROC analysis ,the Area under the ROC curve(AUC) values for SVM, BPN classifiers in wavelet based method, SGLDM method are 0.96,0.91.0.90,0.89. Results show that SVM classifier with wavelet based statistical texture features were yielding better results compared to the statistical texture features extracted directly from the image without applying wavelet transform. This justifies the choice of using wavelet transform. Use of large data bases is expected to improve the system performance and ensure the repeatability of the resulted performance. This approach has potential for further development because of this simplicity that will motivate to classify the types of tumors. The developed classification system is expected to provide valuable diagnosis for the physicians.

**Authors**

**A. Padma** received her B.E in Computer science and Engineering from Madurai Kama raj University ,Tamilnadu ,India in 1990. She received her M.E in Computer science and Engineering from Madurai Kamaraj University ,Tamilnadu, India in 1997. She is currently pursuing P.hd under the guidance of Dr.(Mrs.).R.Sukanesh in Medical Imaging at Anna university, Trichy. She is working as Asst Professor in Department of information Technology, Velammal college of Engineering and Technology,Tamilnadu, India. Her area Of interest includes image processing, Neural Networks, Genetic algorithm. She is a Life member of ISTE,CSI. Chapters.

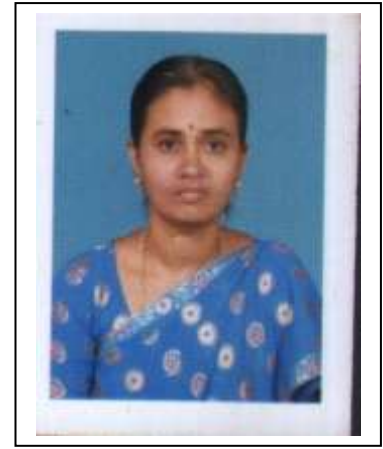

**Dr.(Mrs.).R.Sukanesh**, Professor in Bio medical Engineering has received her B.E from Government college of Engineering and Technology, Coimbatore in 1982. She obtained her M.E from PSG college of Engineering and Technology, Coimbatore in 1985 and Doctoral degree in Bio medical Engineering from Madurai Kama raj university , Madurai in 1999. Since 1985, she is working as a faculty in the Department of Electronics and Communication Engineering, Madurai and presently she is the Professor of ECE, and heads the Medical Electronics division in the same college. Her main research areas include Neural Networks, Bio signal processing and Mobile communication. She is guiding twelve P.hD candidates in these areas. She has published several papers in National, International journals and also published around eighty papers in National and International conferences both in India and Abroad. She is a reviewer of international Journal of Biomedical Sciences and International Journal of signal Processing. She is a life member of Biomedical Society of India.